\documentclass[conference]{IEEEtran}
\IEEEoverridecommandlockouts
\usepackage{cite}
\usepackage{amsmath,amssymb,amsfonts}
\usepackage{algorithmic}
\usepackage{graphicx}
\usepackage{textcomp}
\usepackage{xcolor}
\usepackage{booktabs}
\usepackage{subcaption}
\usepackage{multirow}
\usepackage{hyperref}
\hypersetup{
    colorlinks=true,
    linkcolor=blue,
    filecolor=magenta,      
    urlcolor=cyan,
    pdftitle={Overleaf Example},
    pdfpagemode=FullScreen,
    }

\begin{document}

\title{GCF: Graph Convolutional Networks for Facial Expression Recognition}

\author{\IEEEauthorblockN{ Hozaifa Kassab, Mohamed Bahaa, Ali Hamdi}
\IEEEauthorblockA{\textit{MSA University} \\
 Giza, Egypt \\
\{hozaifa.fadl, mohamed.bahaa4, ahamdi\}@msa.edu.eg }}

\maketitle

\begin{abstract}
Facial Expression Recognition (FER) is vital for understanding interpersonal communication. However, existing classification methods often face challenges such as vulnerability to noise, imbalanced datasets, overfitting, and generalization issues. In this paper, we propose GCF, a novel approach that utilizes Graph Convolutional Networks for FER. GCF integrates Convolutional Neural Networks (CNNs) for feature extraction, using either custom architectures or pretrained models. The extracted visual features are then represented on a graph, enhancing local CNN features with global features via a Graph Convolutional Neural Network layer. We evaluate GCF on benchmark datasets including CK+, JAFFE, and FERG. The results show that GCF significantly improves performance over state-of-the-art methods. For example, GCF enhances the accuracy of ResNet18 from 92\% to 98\% on CK+, from 66\% to 89\% on JAFFE, and from 94\% to 100\% on FERG. Similarly, GCF improves the accuracy of VGG16 from 89\% to 97\% on CK+, from 72\% to 92\% on JAFFE, and from 96\% to 99.49\% on FERG. We provide a comprehensive analysis of our approach, demonstrating its effectiveness in capturing nuanced facial expressions. By integrating graph convolutions with CNNs, GCF significantly advances FER, offering improved accuracy and robustness in real-world applications. The code is available at: \url{https://github.com/4qlaa7/GCF}
\end{abstract}


\section{Introduction}
Facial Expression Recognition (FER) is a key area in computational psychology and human-computer interaction, focusing on interpreting the emotional language conveyed through facial cues. FER involves the automated detection and classification of human emotions, encompassing a range of feelings such as anger, disgust, fear, happiness, sadness, and surprise. This field has significant applications in areas such as social robotics,
healthcare\cite{bisogni2022impact}, and security\cite{liang2020fine}. However, existing FER classification methods face challenges such as noise from environmental factors, imbalanced datasets, and issues with overfitting and generalization\cite{MANALU2024200339}. Despite these obstacles, efforts to improve FER continue, motivated by the potential to enhance human-machine interaction, support assistive technologies, and deepen our understanding of human behavior and cognition.

Deep learning has significantly advanced various domains of computer vision \cite{hamdi2021c,hamdi2021marl,hamdi2020drotrack,al2018arabic}, including FER \cite{MATSUGU2003555,Lopes2017FacialER}. CNNs have become the standard for image processing tasks due to their proficiency in capturing spatial hierarchies of features in images. By automating the feature extraction process, CNNs eliminate the need for manual feature engineering and allow models to learn directly from vast amounts of data, improving scalability and performance. However, CNN-based FER requires extensive labeled datasets and substantial computational power, which can be limiting when resources are constrained or when data is scarce. Hybrid models, integrating CNNs with other machine learning or neural network frameworks, have enhanced the robustness and efficiency of FER. For example, combining CNNs with Recurrent Neural Networks (RNNs) has proven effective for video-based FER by capturing dynamic changes in expressions over time. Additionally, using Support Vector Machines (SVMs) as classifiers on features extracted by CNNs has improved decision boundaries for better classification accuracy. We present GCF, a pioneering methodology that integrates CNNs with Graph Convolutional Networks (GCNs) to enhance FER. Our approach leverages the robust feature extraction capabilities of CNNs and the relational modeling strengths of GCNs to analyze distinct facial regions. This synergy provides a comprehensive understanding of both local nuances and global context in facial expressions, enabling nuanced analysis of emotional states. By exploiting the structural information in GCNs, our model navigates the complexities of facial expressions, surpassing the limitations of traditional FER methodologies \cite{inproceedings}. We evaluate our CNN-GCN model using benchmark FER datasets, including CK+, FERG, and RAF, which cover diverse expression scenarios. Our results show that the proposed model exceeds state-of-the-art performance, outperforming CNN-only and other hybrid systems. The paper is structured as follows: We first review related works in FER, then describe our framework, followed by dataset descriptions, experiments and results analysis, discussion, and conclusions with implications for future research.

\begin{figure*}
    \centering
    \includegraphics[width=1\linewidth]{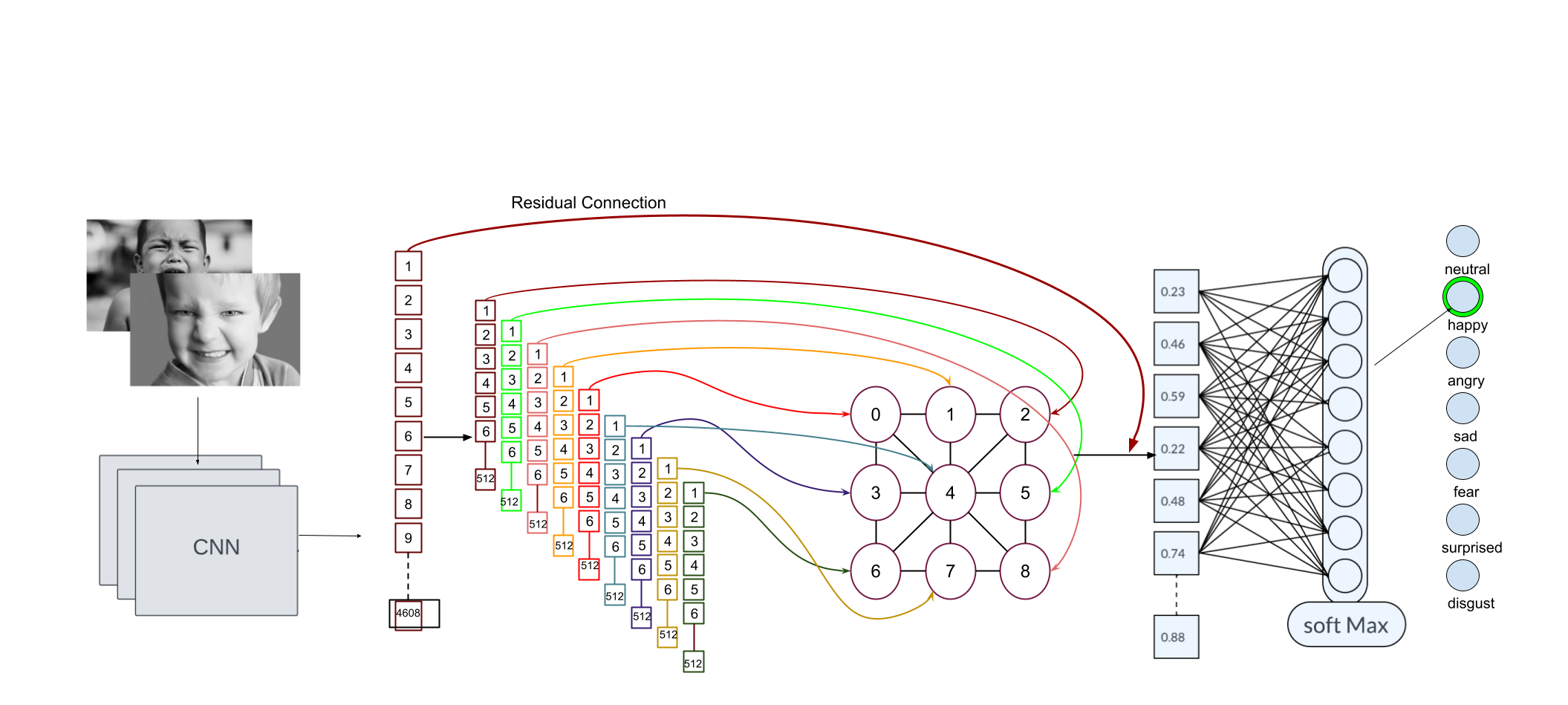}
    \caption{GCF, the proposed model design for EFR using Graph Convolutional Neural Networks.}\vspace{-5mm}
    \label{fig:enter-label}
\end{figure*}

\section{Related Work}
This section reviews the progression of methodologies employed in FER, highlighting the evolution from CNNs to hybrid models integrating CNNs with other computational techniques, and highlighting the need graph-based methods.

CNNs have been widely employed across various computer vision tasks, including Facial Expression Recognition (FER). Multiple studies demonstrated the robustness of CNNs to changes in face location and scale variations, outperforming Multilayer Perceptrons (MLPs) particularly in scenarios involving unseen face poses \cite{MATSUGU2003555}. Another study \cite{meng2019frame} utilized CNNs to tackle challenges such as subject independence, as well as translation, rotation, and scale invariance in facial expression recognition.
Early applications of CNNs in FER often involved training models from scratch. For instance,  a custom CNN architecture  significantly outperformed traditional machine learning approaches by automatically learning features directly from facial images \cite{Lopes2017FacialER}. The advent of pre-trained models brought about a paradigm shift, with researchers leveraging models trained on large, diverse datasets to improve FER accuracy. The work in \cite{donahue2013decaf} demonstrated that features extracted from deep layers of pre-trained CNNs could be repurposed for various vision tasks, including emotion recognition. This approach not only improved the accuracy but also reduced the training time and resource consumption. VGG model pre-trained on ImageNet for FER, achieving robust performance across multiple datasets.

Incorporating Graph Convolutional Neural Networks (GCNs) alongside Convolutional Neural Networks (CNNs) offers a promising avenue for advancing Facial Expression Recognition (FER). By fusing CNNs' robust feature extraction capabilities with GCNs' capacity to model relational data among facial regions, this approach enables a more comprehensive understanding of both local nuances and global context within facial expressions. The synergy between CNNs and GCNs addresses the subtleties and complexities inherent in facial expressions, leading to improved FER performance compared to traditional methods\cite{DBLP:journals/corr/abs-2010-13386},\cite{deng2024multiscale}.

Recent advancements in Facial Expression Recognition (FER) have significantly benefited from hybrid models that integrate Convolutional Neural Networks (CNNs) with Graph Convolutional Networks (GCNs), effectively leveraging the strengths of both methodologies. The MER-GCN framework enhances micro-expression recognition by modeling relational dependencies among facial action units (AUs) using GCNs, with spatial-temporal features extracted by a 3D ConvNet and processed through a stacked GCN, resulting in notable improvements in recognizing subtle and transient expressions\cite{DBLP:journals/corr/abs-2004-08915}. However, the reliance on co-occurrence probabilities to form the adjacency matrix can be limited by the quality and representativeness of the training data, potentially leading to biased or inaccurate relationships being modeled. Another approach constructs an undirected graph from facial images using fixed and random points, where GCNs handle non-Euclidean data structures, capturing complex relationships between facial features to enhance recognition accuracy\cite{8866311}. Despite its innovative approach, the method's effectiveness heavily depends on the initial graph construction, which may not always capture the most relevant features for FER. Additionally, a novel method for video-based FER employs a GCN to represent each video frame as a node, using a dynamic adjacency matrix to model dependencies and refine features through a Bidirectional Long Short-Term Memory (BiLSTM) network, thus capturing both spatial and temporal variations effectively\cite{DBLP:journals/corr/abs-2010-13386}. While this approach addresses dynamic expression variations well, it can be computationally intensive, requiring substantial resources for real-time applications. These models demonstrate the powerful synergy of CNNs and GCNs in advancing FER, but they also highlight the challenges in balancing complexity, data dependency, and computational efficiency.

\section{The Proposed Framework}
In this section, we provide a comprehensive overview of our proposed methodology for facial expression recognition using the GCF model, inspired by \cite{DBLP:journals/corr/abs-2007-15444}
\cite{DBLP:journals/corr/abs-2110-11664}
\cite{DBLP:journals/corr/abs-2110-11551}.

\subsection{CNN for Feature Extraction}
Initially, a CNN is employed to extract relevant features from facial images, utilizing either pre-trained models like VGG or ResNet, or a custom-designed architecture. The CNN processes input images \( X \) to produce a feature representation \( F_{\text{CNN}}(X) \) that captures various facial characteristics, such as edges, textures, and patterns associated with different expressions.The output of the CNN, \( F_{\text{CNN}}(X) \), is then sliced into nine vectors:
\[F_{\text{CNN}}(X) = [f_1, f_2, \ldots, f_9]\]
where \( f_i \) represents the feature vector corresponding to the \( i \)-th facial region. Pre-trained CNNs or other models are favoured for feature extraction in machine learning and computer vision due to their ability to learn rich hierarchical representations. These models, trained on large datasets like ImageNet, allow practitioners to benefit from learned features without needing to train entire models from scratch, significantly reducing computational costs and training time. Fine-tuning these models on task-specific datasets enhances their effectiveness, tailoring representations to specific tasks. This strategy boosts performance, improves generalization, and accelerates the development of state-of-the-art methodologies.

\subsection{Graph Convolutional Neural Network (GCN)}

Each of the nine vectors \( f_i \) is fed into a node within a Graph Convolutional Neural Network (GCN). GCN is a type of neural network tailored for analyzing and processing graph-structured data. Its mechanism revolves around adapting convolutional operations from traditional image-based Convolutional Neural Networks (CNNs) to operate directly on graph structures.

Let \( A \) be the adjacency matrix representing the graph structure, and \( H^{(0)} = [f_1, f_2, \ldots, f_9] \) be the initial node features. A single GCN layer can be represented as:
\[
H^{(l+1)} = \sigma \left( D^{-\frac{1}{2}} A D^{-\frac{1}{2}} H^{(l)} W^{(l)} \right)
\]
where:
\begin{itemize}
    \item \( H^{(l)} \) is the feature matrix at layer \( l \),
    \item \( W^{(l)} \) is the trainable weight matrix at layer \( l \),
    \item \( D \) is the degree matrix of \( A \),
    \item \( \sigma \) is an activation function, typically ReLU.
\end{itemize}

This aggregation process enables the network to extract hierarchical representations of the graph, where deeper layers capture increasingly abstract features.

\subsection{Feature Concatenation and Classification}

After processing through the GCN, the output features from the nine nodes are aggregated into a single feature vector \( H_{\text{GCN}} \). This vector is then concatenated with the initial output feature vector from the CNN before slicing. Let \( F_{\text{CNN-initial}} \) be the initial output feature vector from the CNN before slicing. The concatenated feature vector is:
\[
F_{\text{concat}} = [F_{\text{CNN-initial}}, H_{\text{GCN}}]
\]

This concatenation creates a comprehensive feature representation that includes both detailed local features and broader contextual information. The final feature vector \( F_{\text{concat}} \) is then fed into a classifier (e.g., a fully connected layer followed by a softmax function) to predict the facial expression:
\[
y = \text{softmax}(W_{\text{fc}} F_{\text{concat}} + b_{\text{fc}})
\]
where \( W_{\text{fc}} \) and \( b_{\text{fc}} \) are the weights and biases of the fully connected layer. By leveraging the strengths of both CNN and GCN, our proposed framework efficiently captures intricate facial features and their relationships, enhancing the accuracy and robustness of facial expression recognition systems across various datasets. The effectiveness of our proposed methodology, which combines feature extraction via a CNN with subsequent processing by a Graph Convolutional Network (GCN), represents a significant advancement in facial expression recognition. By first utilizing a CNN, either custom-designed or pre-trained, our approach efficiently captures intricate facial features at multiple scales. These features are then fed into the GCN, which excels in handling structured data, allowing for the modeling of relationships between various facial regions through its graph-based nature. This dual-stage processing not only enhances the representational power of the system but also significantly improves the accuracy and robustness of expression recognition across diverse datasets. Through rigorous experimentation on benchmark datasets such as CK+, JAFFE, and FERG, our proposed model demonstrates superior performance, achieving higher accuracy rates compared to state-of-the-art methods. The results validate the efficacy of integrating CNNs with GCNs, offering a promising direction for future research in the domain of facial expression recognition.

\begin{figure}
   \centering
   \includegraphics[width=0.47\textwidth]{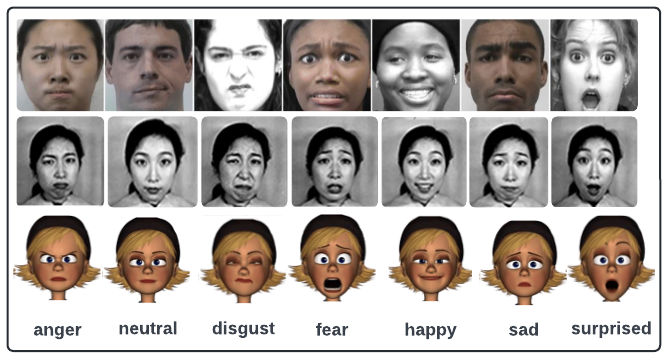}
   \caption{Basic Emotions From CK+, JAFEE, and, FERG dataset}\vspace{-5mm}
\end{figure}

\section{Datasets}
In our study, we conduct a comprehensive experimental analysis of our proposed model using several well-known facial expression recognition datasets: the Extended Cohn-Kanade (CK+), the Facial Expression Research Group Database (FERG), and the Real-world Affective Faces Database (RAF). Below is a detailed overview of these databases, which are crucial for evaluating the effectiveness of our approach. \\
\textbf{Cohn-Kanade The Extended Cohn-Kanade dataset(CK+)}, commonly referred to as CK+, is a widely used public dataset in the realm of action unit and emotion-specified facial expression recognition. This dataset encompasses $593$ sequences from $123$ subjects, featuring both posed and spontaneous expressions. In most research, including ours, the last frame of these sequences is extracted and utilized for static facial expression recognition, capturing the peak of the emotional expression. \textbf{Facial Expression Research Group Database (FERG)} The FERG database consists of $55,767$ annotated images representing six stylized characters designed using MAYA, a 3D animation software. These characters are depicted in a range of seven different expressions, providing a unique challenge in recognizing facial expressions from cartoon-style visuals. This dataset allows us to test the adaptability and performance of our model on non-realistic, stylized facial expressions, which is essential for applications in animated films and video games.\textbf{Japanese Female Facial Expression(JAFFE)}
The JAFFE dataset consist of 213 facial images of different facial expressions, sourced from 10 distinct Japanese female subjects. Each participant was instructed to convey seven facial expressions, encompassing six fundamental emotional states along with a neutral expression. These images were meticulously annotated with mean semantic ratings for each emotional expression, as assessed by a panel of 60 annotators\cite{lyons_2021_5147170},\cite{michael_j_lyons_2020_4029680}.

\section{Experimental Design and Results}
In this section, we present and discuss the experimental results obtained from the evaluation of the proposed Graph Convolutional Network-based Facial Expression Recognition (GCF) model. The results are compared against several state-of-the-art approaches using the aforementioned benchmark datasets. The methods included in the comparison represent a range of traditional and deep learning-based approaches for facial expression recognition. We split our dataset into 80\% for training and validation and 20\% for testing.

We designed our extensive experimental work as follows:
\begin{itemize}
    \item Enhancing the state-of-the-art CNN pre-trained models using the proposed GCF.
    \item Benchmarking the proposed model against the state-of-the-art FER models.
    \item Varying the GCF design in an ablation study.
\end{itemize}

\subsection{Enhancing the State-of-the-art Models using GCF}
Table \ref{tab:JAFFE} compares the performance of the GCF method against baseline and state-of-the-art convolutional neural network (CNN) models on the JAFFE dataset. The results highlight the effectiveness of our approach in achieving higher accuracy rates across various architectures. In each comparison, the GCF method outperformed the baseline CNN models. For instance, the GCF method achieved an accuracy rate of 95\% with VGG16, improving upon the CNN baseline accuracy of 92\%, while for VGG19, the GCF method achieved 92\% accuracy, significantly outperforming the CNN baseline of 72\%. Similarly, with ResNet18, the GCF method reached an accuracy of 89\%, a notable improvement from the baseline CNN accuracy of 66\%. In more complex models like EfficientNetB0, the GCF method achieved 90\% accuracy, up from the CNN baseline of 87\%, while with InceptionV3, the GCF method achieved 99.2\%, surpassing the CNN baseline of 97\%. The results highlight the effectiveness of utilizing graph-based representations for facial expression recognition, as the GCF approach consistently achieves higher accuracy rates, even in highly optimized models like EfficientNetB0 and complex architectures like InceptionV3 and DenseNet. This consistent enhancement, with improvements ranging from a few percentage points to over 20\% in certain cases, underscores the potential of graph convolutional networks in advancing the field of facial expression recognition.

\begin{table}[h]
\centering
\begin{tabular}{lcc}
    \toprule
    \textbf{Models} & \textbf{CNN} & \textbf{GCF (Ours)} \\ \midrule
    VGG16 & 92\% & 95\% \\
    VGG19 & 83\% & 92\% \\
    RESNET18 & 70\% & 89\% \\
    RESNET34 & 82\% & 87\% \\
    RESNET50 & 77.3\% & 83\% \\
    EfficientNetB0 & 87\% & 90\% \\
    InceptionV3 & 97\% & 99.2\% \\
    densenet121 & 91.2\% & 98\% \\
    densenet169 & 83\% & 89.2\% \\
    mobilenetv2 & 97\% & 100\% \\ \bottomrule
\end{tabular}
\caption{FER benchmarck on the JAFFE Dataset.}
\label{tab:JAFFE}
\end{table}

The performance evaluation on the CK+ dataset, as presented in Table \ref{tab:CK+}, highlights the superior accuracy achieved by the GCF method compared to baseline convolutional neural network (CNN) models. The GCF method consistently outperformed the CNN counterparts across various architectures. For the VGG models, GCF achieved 97\% and 98\% accuracy for VGG16 and VGG19, respectively, marking significant improvements of 8\% and 5\% over the baseline. Similarly, with the ResNet models, the GCF method improved accuracy rates to 98\%, 96\%, and 94\% for ResNet18, ResNet34, and ResNet50, respectively, outperforming the baselines by 6\%, 3\%, and 7\%. The GCF method also achieved a perfect accuracy of 100\% with EfficientNetB0, surpassing the CNN baseline of 98\%. In the case of InceptionV3 and DenseNet121, the GCF method achieved 99.2\% and 100\%, respectively, slightly improving upon the CNN baselines. Furthermore, with MobileNetV2, the GCF method attained 98\%, surpassing the CNN baseline by 3\%. These consistent enhancements across diverse architectures emphasize the potential of our graph-based approach for FER, solidifying its effectiveness on the CK+ dataset and showcasing the advantages of graph convolutional networks in this domain.

\begin{table}[h]
\centering
\begin{tabular}{lcc}
\toprule
\textbf{Models} & \textbf{CNN} & \textbf{GCF (Ours)} \\ \midrule
VGG16           & 89\% & 97\%   \\
VGG19           & 93\% & 98\%   \\
RESNET18        & 92\% & 98\%   \\
RESNET34        & 93\% & 96\%   \\
RESNET50        & 87\% & 94\%   \\
EfficientNetB0  & 98\% & 100\%  \\
InceptionV3     & 97\% & 99.2\% \\
densenet121     & 97\% & 100\%  \\
densenet169     & 97\% & 98\%   \\
mobilenetv2     & 95\% & 98\%   \\ \bottomrule
\end{tabular}
\caption{Performance comparison on CK+ dataset}
\label{tab:CK+}
\end{table}

The performance evaluation on the FERG dataset, as presented in Table \ref{tab:FERG}, underscores the robust accuracy achieved by GCF method compared to baseline convolutional neural network (CNN) models. Across different architectures, the GCF method consistently outperformed the CNN counterparts. For the VGG models, GCF achieved 99.6\% and 99.4\% accuracy for VGG16 and VGG19, respectively, representing improvements of 1.8\% and 3.3\%. The ResNet models also benefited from the GCF approach, with ResNet18 achieving a perfect 100\% accuracy, an improvement of 5.6\%, and ResNet50 achieving 99.7\%, a slight improvement over the baseline. Even with more efficient architectures, such as EfficientNetB0, the GCF method achieved 99.97\%, surpassing the baseline accuracy of 98\%. The InceptionV3 and DenseNet architectures also saw improvements with GCF, achieving up to 96\% and 98.7\% accuracy, respectively, outperforming their CNN baselines. Finally, with MobileNetV2, the GCF method achieved 99\%, surpassing the CNN baseline by 3\%. These results not only show the impact of the GCF upon diverse architectures but also highlight the adaptability and superior performance of graph convolutional networks for facial expression recognition on the FERG dataset.

\begin{table}[h]
\centering
\begin{tabular}{lcc}
    \toprule
    \textbf{Models} & \textbf{CNN} & \textbf{GCF (Ours)} \\ \midrule
    VGG16           & 97.8\% & 99.6\% \\
    VGG19           & 96.1\% & 99.4\% \\
    RESNET18        & 94.4\% & 100\% \\
    RESNET34        & 93\%   & 96\% \\
    RESNET50        & 98\%   & 99.7\% \\
    EfficientNetB0  & 98\%   & 99.97\% \\
    InceptionV3     & 93\%   & 96\% \\
    densenet121     & 97\%   & 98.7\% \\
    densenet169     & 96\%   & 98\% \\
    mobilenetv2     & 96\%   & 99\% \\ \bottomrule
\end{tabular}
\caption{Performance comparison on FERG dataset.}\vspace{-2mm}
\label{tab:FERG}
\end{table}

\subsection{Benchmark with the State-of-the-art Models}
The comprehensive performance comparison across the CK+, JAFFE, and FERG datasets, as presented in Table \ref{tab:accuracy_comparison}, highlights the robust accuracy achieved by the GCF method compared to several state-of-the-art models. In this benchmark with the state-of-the-art models, our GCF method consistently outperformed prominent approaches. On the CK+ dataset, our GCF method achieved a perfect accuracy rate of 100\%, outperforming notable models such as DeepEmotion (98\%), Nonlinear Evaluation on SL + SSL Puzzling (98.23\%), FAN (99.7\%), and ViT + SE (99.8\%). Similarly, on the JAFFE dataset, the GCF method also achieved a perfect accuracy rate of 100\%, surpassing strong baselines like DeepEmotion (92.8\%), ViT (94.83\%), ARBEx (96.67\%), and TLE (99.52\%). The performance trend continued on the FERG dataset, where the GCF method achieved an impressive accuracy rate of 99.98\%, outperforming DeepEmotion (99.3\%) and ARBEx (98.18\%). These results underscore the efficacy and robustness of our GCF method across diverse facial expression recognition datasets, highlighting its superior performance over both traditional and cutting-edge deep learning approaches. The consistently high accuracy rates achieved across different datasets demonstrate the adaptability and effectiveness of the GCF method, making it a promising approach for facial expression recognition tasks.

\begin{table*}[h]
\centering
\begin{tabular}{lcc|lcc|lcc}
    \toprule
    \multicolumn{3}{c}{\textbf{CK+}} & \multicolumn{3}{|c|}{\textbf{JAFFE}} & \multicolumn{3}{c}{\textbf{FERG}} \\ \midrule
    \textbf{Method} & \textbf{Accuracy} & & \textbf{Method} & \textbf{Accuracy} & & \textbf{Method} & \textbf{Accuracy} & \\ \midrule
    DeepEmotion\cite{minaee2019deepemotion} & 98\% & & DeepEmotion\cite{minaee2019deepemotion} & 92.8\% & & DeepEmotion\cite{minaee2019deepemotion} & 99.3\% & \\ 
    Nonlinear eval on SL + SSL puzzling  & 98.23\% & & ViT\cite{DBLP:journals/corr/abs-2107-03107} & 94.83\% & & ARBEx\cite{wasi2023arbex} & 98.18\% & \\ 
    FAN \cite{meng2019frame}  & 99.7\% & & ARBEx\cite{wasi2023arbex} & 96.67\% & &  & \\ 
    ViT + SE \cite{DBLP:journals/corr/abs-2107-03107} & 99.8\% & & TLE\cite{electronics10091036} & 99.52\% & \\ \midrule
    GCF (Ours) & \textbf{100\%} & & GCF (Ours) & \textbf{100\%} & & GCF (Ours) & \textbf{99.98\%}\\ \bottomrule
\end{tabular}
\caption{Comparison of accuracy across CK+, JAFFE, and FERG datasets.}
\label{tab:accuracy_comparison}
\end{table*}

\begin{figure}
    \centering
    \includegraphics[width=0.48\textwidth]{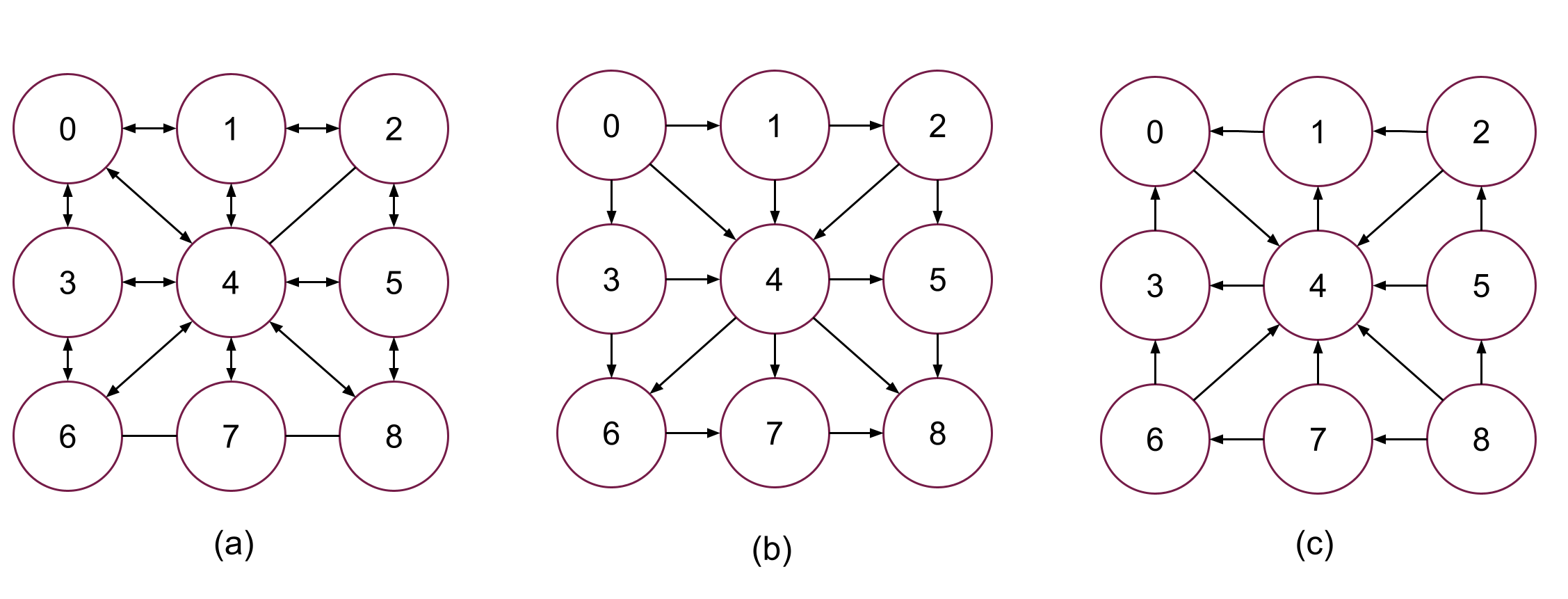}
    \caption{GCF versions.}\vspace{-5mm}
    \label{GCF-vs}
\end{figure}

\subsection{Ablation Study}
The focus of this research is on the movements of the graph approach and its significance when the direction of edges in the graph is taken into account. Three models are identified: GCF-V1 with bidirectional edges, GCF-V2 with cross-linked edges from left to right, and GCF-V3 with cross-linked edges from right to left. The results summarize the accuracy rates of facial expression recognition models across three key datasets: JAFFE, CK+, and FERG. The study considers pretrained models including VGG16, ResNet18, and InceptionV3 and utilizes them for the three versions of the Graph Convolutional Facial Expression (GCF) distorted dataset. This work is proof of the power of certain neural network architectures for recognizing diverse facial patterns indicated in several datasets and versions of algorithms. GCF variations between directional arrangements can be used to acquire the quantitative effect of different organizational schemes on model performance.

\begin{table}[h]
\centering
\begin{tabular}{lccccccccccccc}
\toprule
\textbf{Dataset} & \textbf{Models} & \textbf{GCF-V1} & \textbf{GCF-V2} & \textbf{GCF-V3} \\ \midrule
\multirow{10}{*}{JAFFE} & VGG16 & 95\% & 93\% & 86\% \\
 & VGG19 & 92\% & 90\% & 83\% \\
 & RESNET18 & 89\% & 63\% & 83\% \\
 & RESNET34 & 87\% & 86\% & 82\% \\
 & RESNET50 & 83\% & 77\% & 80\% \\
 & EfficientNetB0 & 90\% & 78\% & 86\% \\
 & InceptionV3 & 99.2\% & 91\% & 88\% \\
 & densenet121 & 98\% & 90\% & 88\% \\
 & densenet169 & 89.2\% & 83\% & 91\% \\
 & mobilenetv2 & 100\% & 80\% & 83\% \\ \midrule
\multirow{10}{*}{CK+} & VGG16 & 97\% & 98\% & 99\% \\
 & VGG19 & 98\% & 99.4\% & 96\% \\
 & RESNET18 & 98\% & 99\% & 99.4\% \\
 & RESNET34 & 96\% & 98\% & 98.6\% \\
 & RESNET50 & 94\% & 99.3\% & 94\% \\
 & EfficientNetB0 & 100\% & 96\% & 95.8\% \\
 & InceptionV3 & 99.2\% & 97\% & 99.1\% \\
 & densenet121 & 100\% & 97.7\% & 98.4\% \\
 & densenet169 & 98\% & 95\% & 99.6\% \\
 & mobilenetv2 & 98\% & 97\% & 96.7\% \\ \midrule
\multirow{10}{*}{FERG} & VGG16 & 99.6\% & 97\% & 98\% \\
 & VGG19 & 99.4\% & 96\% & 97\% \\
 & RESNET18 & 100\% & 95.6\% & 96\% \\
 & RESNET34 & 96\% & 92\% & 94\% \\
 & RESNET50 & 99.7\% & 98\% & 96\% \\
 & EfficientNetB0 & 99.97\% & 98\% & 98.8\% \\
 & InceptionV3 & 96\% & 94\% & 94\% \\
 & densenet121 & 98.7\% & 97\% & 97.5\% \\
 & densenet169 & 98\% & 96\% & 93\% \\
 & mobilenetv2 & 99\% & 97\% & 95\% \\ \bottomrule
\end{tabular}
\caption{Benchmark different GCF versions
}
\label{tab:ablation}
\end{table}
\section{Conclusion}
In this paper, we introduced a novel approach to Facial Expression Recognition (FER) by combining Convolutional Neural Networks (CNNs) and Graph Convolutional Networks (GCNs). GCF model leverages the feature extraction capabilities of CNNs and the relational modeling strengths of GCNs to capture both local and global features of facial expressions. Our experiments on benchmark datasets, including CK+, JAFFE, and FERG, demonstrated significant improvements in accuracy, achieving up to 100\% accuracy on the FERG dataset. These results highlight that The GCF framework represents a promising direction for future research in facial expression recognition, providing a deeper understanding of human emotions.
\section{Acknowledgment}
We extend our heartfelt gratitude to \textit{AiTech for Artificial Intelligence \& Software Development} (\url{https://aitech.net.au}) for providing the computational resources essential for our experiments. Their support has been crucial to the successful completion of this research.

\bibliographystyle{plain}
\bibliography{main}

\end{document}